%% file: ICORR_2025.tex
\documentclass[letterpaper, 10 pt, conference]{ieeeconf}
\IEEEoverridecommandlockouts
\usepackage{gensymb}
\usepackage{graphicx}
\usepackage{amsmath}
\usepackage{cite}
\usepackage{booktabs}  
\usepackage{makecell}   
\usepackage[table]{xcolor}
\usepackage{colortbl}

\title{\LARGE \bf
Personalization of Wearable Sensor-Based Joint Kinematic Estimation Using Computer Vision for Hip Exoskeleton Applications}

\author{Changseob Song$^{1}$*, Bogdan Ivanyuk-Skulskyi$^{2}$*, Adrian Krieger$^{1}$, Kaitao Luo$^{1}$, and Inseung Kang$^{1}$
\thanks{This work was supported in part by the Kwanjeong Educational Foundation. Changseob Song and Bogdan Ivanyuk-Skulskiy contributed equally to this work. Corresponding author: Changseob Song, {\tt\footnotesize changseob@cmu.edu}.}
\thanks{$^{1}$C. Song, A. Krieger, K. Luo, and I. Kang are with the Department of Mechanical Engineering, Carnegie Mellon University, Pittsburgh, PA 15213,
USA.}
\thanks{$^{2}$B. Ivanyuk-Skulskyi is with the  CRIStAL, University of Lille, Lille, 59000, France and Cyclope.ai, Nanterre, 92000, France.}
}

\begin{document}

\maketitle
\begin{abstract}

Accurate lower-limb joint kinematic estimation is critical for applications such as patient monitoring, rehabilitation, and exoskeleton control. While previous studies have employed wearable sensor-based deep learning (DL) models for estimating joint kinematics, these methods often require extensive new datasets to adapt to unseen gait patterns. Meanwhile, researchers in computer vision have advanced human pose estimation models, which are easy to deploy and capable of real-time inference. However, such models are infeasible in scenarios where cameras cannot be used. To address these limitations, we propose a computer vision-based DL adaptation framework for real-time joint kinematic estimation. This framework requires only a small dataset (i.e., 1–2 gait cycles) and does not depend on professional motion capture setups. Using transfer learning, we adapted our temporal convolutional network (TCN) to stiff knee gait data, allowing the model to further reduce root mean square error by 9.7\% and 19.9\% compared to a TCN trained on only able-bodied and stiff knee dataset, respectively. Our framework demonstrated a potential for smartphone camera-trained DL model to estimate real-time joint kinematics across novel users in clinical populations with applications in wearable robots.
    
\end{abstract}


\section{Introduction}

Estimating lower-limb joint kinematics has been an important agenda in patient monitoring, rehabilitation, and exoskeleton control \cite{SHULL201411, BENSON2018124}. Accurate and rapid estimation of joint kinematics allows biomechanics and exoskeleton researchers to better understand gait patterns and leverage this information to assist users more effectively. This is important for quantifying irregular and pathological gait patterns, which helps identify gait characteristics and find out appropriate treatment or assistance strategies \cite{SPAIN2012573, ADKIN2005240}. Hip exoskeletons are particularly important because they support the hip joint, which plays a major role in locomotion energetics. \cite{Sawicki_2009_Spring, Sawicki_2020_Review}. Precise joint kinematic data is essential for controlling hip exoskeletons: it allows real-time synchronization with the user's natural gait and facilitates personalized assistance \cite{For_JBSPC_2022, KIM_Science_2019}.
    
Recent biomechanics and exoskeleton studies have utilized deep learning (DL) models to estimate biomechanical parameters from wearable sensor data, eliminating the need for professional motion capture setups \cite{Shin_ML_kinematic, HERNANDEZ2021185}. Among wearable sensors, inertial measurement units (IMUs) are particularly popular, as they capture kinematic information of limb segments, such as acceleration and angular velocity \cite{Shin_ML_kinematic, HERNANDEZ2021185}. DL models leverage this data to accurately predict users’ joint kinematics in real time. However, training these models still requires motion capture systems to provide ground-truth labels.
    
Although many studies have used motion capture systems to extract body kinematics, motion capture systems are expensive, hard to setup, and cannot be easily moved to different places \cite{OpenCap}. Particularly, marker-based motion capture is cumbersome, requiring over 10 minutes to attach the markers, and it is accessible only to skilled operators who can accurately locate anatomical key points. Additionally, marker-based methods yield less repeatable results because different experimenters may attach markers in different positions \cite{KANKO2021110422}. At the same time, marker-less motion capture systems involve artificial intelligence (AI)-based algorithms, which are proprietary and require considerable computing resources. Therefore, motion capture systems are only applicable in professional biomechanical facilities and are hard to be used in home settings or remote clinics at small-sized hospitals.

Human pose estimation (HPE) leverages AI models to detect skeletal joints from raw image frames, estimate the human poses, and reconstruct three-dimensional (3D) skeletal models \cite{10.1145/3603618, WANG2021103225}. This approach enables the use of a single monocular camera, such as a personal smartphone, to estimate human poses and build kinematic models \cite{10.1145/3603618}. Using open-source algorithms, joint kinematics can be easily extracted without incurring additional costs or requiring labor-intensive preparation. These tools can efficiently capture gait information from untrained users, providing valuable biomechanical data.
    
Recent studies have explored deploying vision models to estimate joint kinematics. For example, OpenCap, a portable biomechanics system, utilizes minimal smartphone cameras to monitor gait patterns \cite{OpenCap}. Cotton et al. developed a framework to generate a smooth trajectory of joint positions \cite{cotton2023markerlessmotioncapturebiomechanical} and fused IMU data with vision-based motion data to enhance knee joint angle estimation \cite{10719724}. Similarly, Shin et al. established a joint kinematic estimation pipeline using synthetic IMU and video data from extensive datasets \cite{Shin_TBME_2023}. While these studies deployed HPE models to reconstruct 3D joint kinematics, these technologies cannot run in real time for practical uses, especially during outdoor locomotion. Compact vision models capable of real-time operation have been proposed \cite{Choi_2021_CVPR, VNect}, but they are not applicable for daily outdoor ambulations and exhibit less accurate estimation. Access to a reliable, real-time joint kinematic estimation mechanism that can adapt effectively across diverse users would significantly enhance applications requiring real-time functionality, such as robotic exoskeleton control \cite{For_JBSPC_2022, 9364364}.

Here, we propose an adaptive joint kinematic estimation framework that leverages computer vision and DL models to accommodate users with irregular gait patterns. We hypothesize that computer vision-based HPE models can serve as a new ground truth for training DL models for estimating irregular gait patterns with minimal experimental settings. This can be achieved by using an open-source HPE pipeline with a monocular video input. By capturing irregular gait patterns and utilizing this information to train the wearable sensor-based DL models, we expect significant improvement in model performance. This approach will enable biomechanics researchers to have valuable insights to monitor each patient's gait patterns. Additionally, our framework can enhance existing exoskeleton control systems by rapidly personalizing the estimation model to accommodate individuals with irregular gait patterns.

\section{Methods}

\subsection{Wearable Sensing Suit}

\input{figures/wearable_suit}

We built a wearable sensing suit that can collect IMU data and estimate joint kinematics in real time. The sensing suit consists of 3 IMUs (ICM-20948, TDK InvenSense, Japan) positioned at the pelvis, left thigh, and right thigh, which are securely fastened by a Velcro strap and a compression band as shown in Fig.~\ref{Sensing_suit}a. We used a portable computer (Model 4B, Raspberry Pi, UK) to both collect IMU data via I2C communication protocol and perform real-time inference, as illustrated in Fig.~\ref{Sensing_suit}b. The portable computer was powered by a portable battery inside of a 3D-printed control box. To synchronize ground-truth labels with timestamps logged by a data acquisition box (Lock Lab, Vicon, USA), the potable computer's software architecture was programmed to receive a 50 Hz trigger from the data acquisition box. Each iteration took less than 20 ms to log the sensor data and perform real-time inference.

\subsection{Computer Vision-Based joint Kinematic Estimation Pipeline}

\input{figures/vision_model_scheme}
\input{figures/setup}
\input{figures/tcn}
We used computer vision models to extract the user's body kinematics. We utilized the MMpose \cite{mmpose2020} library to build the pipeline presented in Fig.~\ref{vision_pipeline}a that includes: (i) object detection; (ii) two-dimensional (2D) keypoint estimation; (iii) 2D-to-3D keypoint uplifting.
       
After obtaining 3D keypoints, we calculated the hip joint angle $\alpha$ by measuring the angle between the vector connecting the pelvis and the spine keypoints, and the vector connecting the hip and the knee keypoints. For the knee joint angle, we calculated the angle $\beta$ between the vector connecting the hip and the knee keypoints, and the vector connecting the knee and the ankle keypoints. We selected one of the six standard RGB cameras (BFS-U3-23S3C-C, Teledyne Vision Solutions, USA) as the video input to our vision model-based pipeline, depicted in Fig.~\ref{Mocap_setup}. Six cameras were also used to measure ground-truth joint angles with a commercial marker-less motion capture system (Theia3D, Theia Markerless Inc., CA).

We used the ViTPose \cite{ViTPose} human body keypoint estimation model that was pre-trained on the COCO dataset \cite{COCO_dataset}, due to its state-of-the-art performance, wide adoption in the field, and robust keypoint predictions. Furthermore, we examined the estimation performance of different ViTPose models and six different camera views depicted in Fig.~\ref{Mocap_setup}. As shown in Table~\ref{table_visionmodel}, we obtained the best joint angle detection result from ViTPose-base in right side camera view, as depicted in Fig.~\ref{vision_pipeline}c. After selecting the vision model and camera view, we applied a Savitzky–Golay filter with a window size of 50 and a fourth-order polynomial to mitigate jittery noise from the vision pipeline. Overall, for further validation experiments, we used YOLOv8 \cite{YOLOv8} as the object detection algorithm, ViTPose-base to detect 2D biological keypoints within the image frames, and Video-Pose3D \cite{VideoPose3D} to reconstruct 3D keypoints from the detected 2D coordinates.

\input{tables/vision_models_camera_table}

\subsection{Temporal Convolutional Network}

For DL models that are used to infer biomechanical information from wearable sensor data, it is essential for DL models to capture temporal features from the input data. While long short-term memory (LSTM) networks \cite{LSTM-Schmidhuber} have been used for processing sequential data, they are slow to train and have substantial memory requirements. Other popular models like recurrent neural networks (RNNs) face a vanishing gradient problem \cite{RNN}, and are challenging to train.
    
Conversely, a temporal convolutional network (TCN) is better suited to handle time-series data because TCNs have dilated convolutional layers, which enable the model structure to parallelize multiple convolutions and learn long-term dependencies in the data \cite{TCN-Use-1}. Therefore, TCNs are faster to train and more memory-efficient than LSTMs or RNNs \cite{TCN-Evaluation} and have been deployed by many previous studies \cite{TCN-Use-Model, TCN-Use-1, TCN-Use-2, TCN-Use-3, TCN-Use-4}.
    
In our TCN implementation, we built five temporal blocks to extract relevant features from the time-sequential 18-channel data of three IMUs. Each temporal block included two convolutional layers and dilation factor. After passing through the temporal blocks, a fully connected layer mapped four joint angle values. We trained the TCN model using an Adam optimizer with a learning rate of 0.001 and a batch size of 32, for 50 epochs with early stopping. Each convolutional layer had 32 channels with a kernel size of 7, followed by ReLU activation function and a dropout rate of 0.1, which was optimized by a hyperparameter sweep. Once the TCN model was trained, we converted it to an open neural network exchange (ONNX) format to minimize inference latency.

\subsection{Experimental Protocol}

We recruited three able-bodied (AB) subjects, one female and two males, with an average age of 25.3 ± 2.1 years, height of 174.7 ± 6.5 cm, and weight of 67.3 ± 9.0 kg. Within each subject, we collected a training dataset in four constant speed conditions (i.e., 0.4, 0.7, 1.0, 1.3 m/s), 1 minute each per trial. Along with collecting IMU data using the sensing suit, we obtained the ground-truth kinematic data using the camera setup shown in Fig.~\ref{Mocap_setup}. Afterwards, to simulate an irregular and asymmetric gait pattern, each AB subject was equipped with a knee brace (Fig.~\ref{Stiff_knee}a) to constrain and stiffen their right knee joint motion, which served as a stiff knee (SK) gait pattern as plotted in Fig.~\ref{Stiff_knee}b. We collected the SK dataset using the same protocol as the AB dataset.
    
After training the TCN model on the AB dataset (AB model), we adapted the baseline model to an AB+SK model via transfer learning using kinematic data extracted from the HPE pipeline. As shown in Fig.~\ref{Stiff_knee}c, we analyzed how SK-to-AB dataset ratio affected estimation error on the SK test dataset. To evaluate model performance, we used the root mean squared error (RMSE) metric, which is particularly advantageous for highlighting outliers by heavily penalizing larger errors. When the SK portion reached 6\% (equivalent to approximately 1 to 2 gait cycles for each speed condition) the AB+SK model's test error (on 10\% of the total dataset) converged to 4.7 ± 0.4 deg RMSE, while the SK model's error remained at 5.9 ± 0.5 deg RMSE. For real-time validation experiments, we utilized the AB+SK model trained with a 6\% SK-to-AB ratio. Similarly, the SK model was trained using the same SK-to-AB ratio, but without leveraging any pre-trained model.

To validate the DL model's estimation performance under various walking speeds and transient conditions with changing speeds, we conducted experiments using a treadmill (FIT5, Bertec, USA) with the speed profile shown in Fig.~\ref{Results}a. Starting from a stationary position, each subject experienced four speeds (1.1, 0.5, 1.2, and 0.6 m/s, sequentially), which were different from the speeds used during TCN model training. The subject also experienced transient intervals with an acceleration or deceleration of 0.5 m/s². The speed profile lasted 185 seconds in total.

\input{figures/stiff_knee}

\section{Results}

To validate real-time estimation performance of the trained TCN model on various subjects, we began by evaluating the AB model's performance on AB subjects as demonstrated in Fig.~\ref{Results}b. For AB model-to-AB subject validation, the hip joint RMSE (3.7 deg) was lower than the knee joint RMSE (7.0 deg). When the AB model was used to estimate the SK subjects' joint kinematics, the hip joint RMSE increased to 5.8 deg, while the knee joint RMSE increased to 10.6 deg. This increase was accompanied by an asymmetry in RMSE values between the right and left legs due to the asymmetric gait patterns, with the right knee joint (stiff knee) exhibiting the highest estimation error among all joints.

For SK model-to-SK subject validation, the average estimation error across all four joints was the highest among all experiments (RMSE of 9.3 deg). Specifically, the hip joint angle estimation RMSE was 6.7 deg, while the knee joint angle estimation RMSE was 11.8 deg. In contrast, using the adapted AB+SK model for SK subjects reduced the hip joint angle RMSE to 4.8 deg and the knee angle estimation RMSE to 10.1 deg, resulting in an average joint angle estimation RMSE of 7.4 deg. This represents an improvement of 9.7\% compared to the AB model-to-SK subject case and 19.9\% compared to the SK model-to-SK subject case. For estimation performances across different speed conditions, estimation performance during high speed intervals (i.e., 1.1 m/s and 1.2 m/s) was the lowest compared to the low speed and transient speed intervals as shown in Fig.~\ref{Results}c.

\input{figures/results}

\section{Discussion}

\input{figures/results_joint}

In this work, we proposed the adaptable joint kinematic estimation framework for irregular gait patterns, particularly using a combination of the computer vision-based HPE algorithm and a TCN model. The vision-trained, adapted AB+SK model exhibited a decreased hip joint angle estimation RMSE of 4.8 deg and the knee joint angle estimation RMSE of 10.1 deg, outperforming the AB model-to-SK subject case by 9.7\% and the SK model-to-SK subject case by 19.9\%. This result corroborates our hypothesis that computer vision models can facilitate wearable sensor-based DL models to adapt to new gait patterns, which could not be learned by a pre-existing DL model that was trained solely for AB gait patterns. This framework has the potential to function as an effective tool for monitoring joint kinematics in real time, especially in scenarios where only limited datasets are available for model training and professional motion capture systems are inaccessible. Furthermore, we anticipate that our system can be deployed to a hip exoskeleton system where DL models can be generalizable to both AB and SK users.

An unexpected outcome was the higher estimation error of the SK model on SK subjects compared to the estimation error of the AB model on the same group, even though the SK gait pattern was different from the AB gait patterns. This could be attributed to our study having a much smaller SK dataset size compared to the AB dataset, which was equivalent to only 1 to 2 gait cycles. Nevertheless, this small dataset portion effectively fine tuned the adapted AB+SK model, resulting in the lowest overall estimation error for SK subjects across all experiments. In addition, knee joint angle estimation errors were consistently higher than the hip joint angle estimation errors. This discrepancy in estimation performance was potentially stemmed from the sensing modality of our suit, which only included pelvis and thigh IMUs. For more precise knee joint estimation (or ankle joints), future work will involve reinforcing our sensing suite with additional IMUs being located at distal lower-limb segments.

The SK model and the AB+SK model also exhibited higher estimation error for the left leg than the right leg in both hip and knee joints. These errors were exhibited due to the model failing to track the hip joint angle around its minimum peak as illustrated in Fig.~\ref{results_joint}. Considering that we utilized the right side view camera among six cameras, occlusion effects likely contributed to these left-side errors, as the vision-extracted kinematic data formed the basis for training both SK and AB+SK models. Considering that we utilized only the right side view camera, the observed errors are likely caused by occlusion effects as outlined in \cite{occlusion}, hindering accurate detection and introducing noise into the joint kinematic data.

In future work, we aim to replace our monocular camera setup with a multiple camera system to implement a triangulation method, mitigating the occlusion effect from single-side views. We will also examine the the estimation accuracy across different vertical camera positions, as well as horizontal camera positions we already tested in (Table~\ref{table_visionmodel}). This analysis will incorporate newly released HPE models, such as Sapiens \cite{Sapiens}, and grounded 3D human body models, such as WHAM \cite{Shin_CVPR_2024_WHAM}. Although one of the reasons we used cameras from the motion capture setup was to easily sync with the ground-truth data, we also plan to build a smartphone camera-based framework for general use. Furthermore, we will expand the size of our AB dataset to enhance the pre-trained model's ability to generalize across diverse gait patterns from different subjects, creating a more robust baseline model that can be further adapted to irregular gait patterns. Lastly, we will implement our wearable sensor-based joint kinematic estimation framework for enhancing robotic hip exoskeleton control in real time.

\section{Conclusion}

We developed a lower-limb joint kinematic estimation framework using a wearable IMU sensor-based TCN model. Additionally, we adapted this model using computer vision-based HPE method. With transfer learning, using a dataset of only 1–2 gait cycles of a new gait pattern, the adapted estimation model achieved improvements of 9.7\% and 19.9\% in RMSE compared to the models trained solely on the AB dataset and the SK dataset, respectively. This pipeline holds promise as a real-time joint kinematic estimation tool that can be quickly adapted to new gait patterns without requiring a specialized setup.

\bibliographystyle{ieeetr}
\bibliography{ICORR_2025}

\end{document}

%% file: figures/wearable_suit.tex
\begin{figure}
    \centering
    \includegraphics[width=0.5\textwidth]{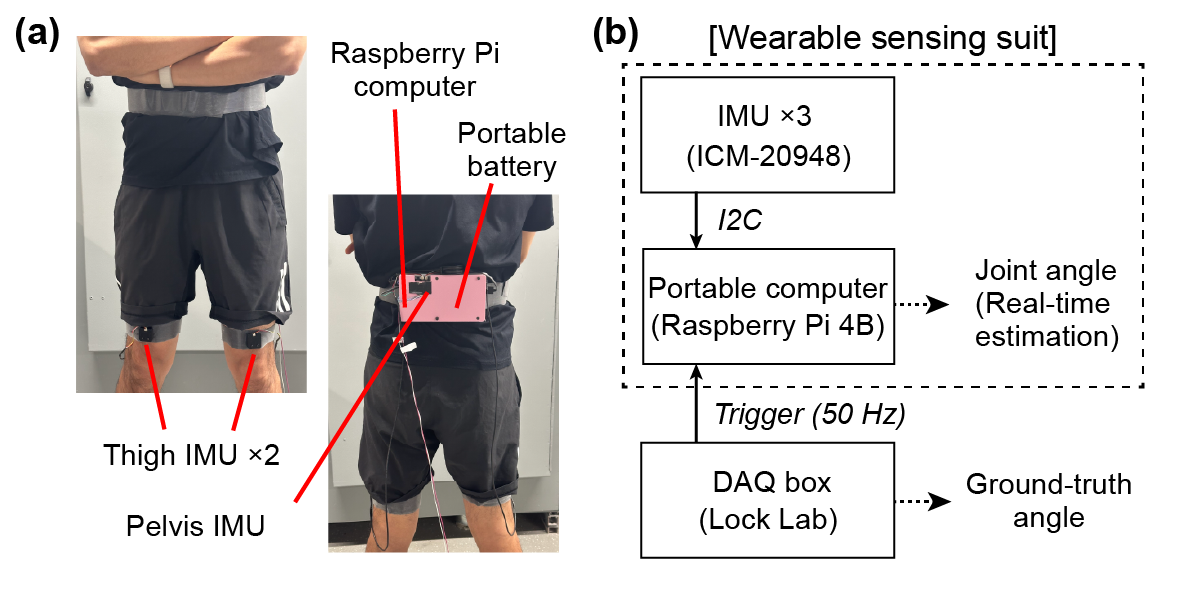}
    \caption{
    Wearable sensing suit for joint angle estimation.
    (a) Components of the sensing suit hardware.
    (b) Data flow within the sensing suit for sensor data logging and real-time inference of joint kinematics.}
    \label{Sensing_suit}
\end{figure}

%% file: figures/vision_model_scheme.tex
\begin{figure}
    \centering
    \includegraphics[width=0.5\textwidth]{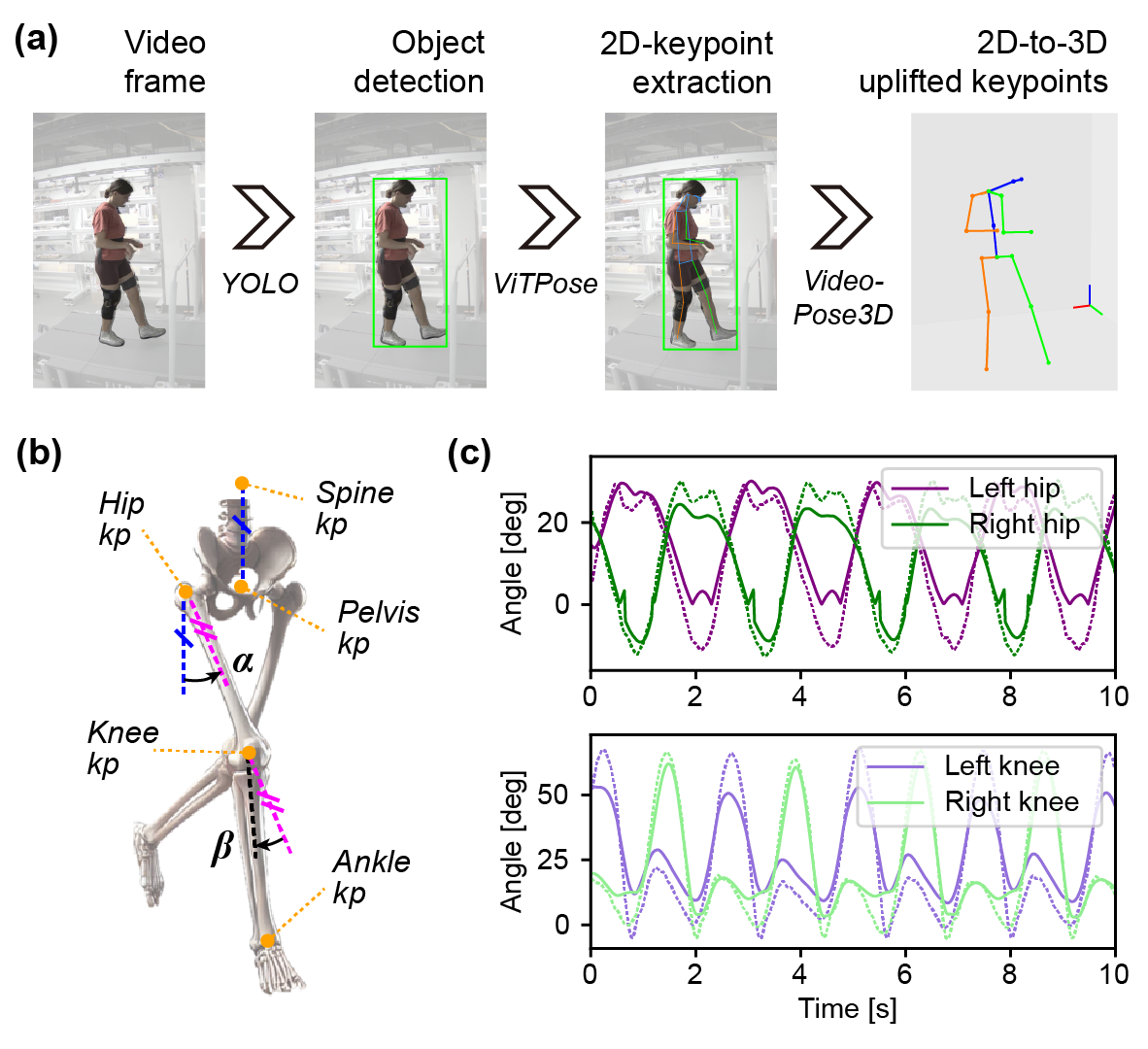}
    \caption{
        Pose estimation pipeline using a human vision model.
        (a) Overall pipeline for extracting 3D joint keypoints from video frames.
        (b) Visual diagram depicting how joint angles are measured from the 3D keypoints.
        (c) Plot of estimated joint angles (solid lines) for an able-bodied subject walking at 1.0 m/s, generated using the pose estimation pipeline, compared to the ground-truth joint angles (dotted lines).
        }
    \label{vision_pipeline}
\end{figure}

%% file: figures/setup.tex
\begin{figure}
        \centering
        \includegraphics[width=0.5\textwidth]{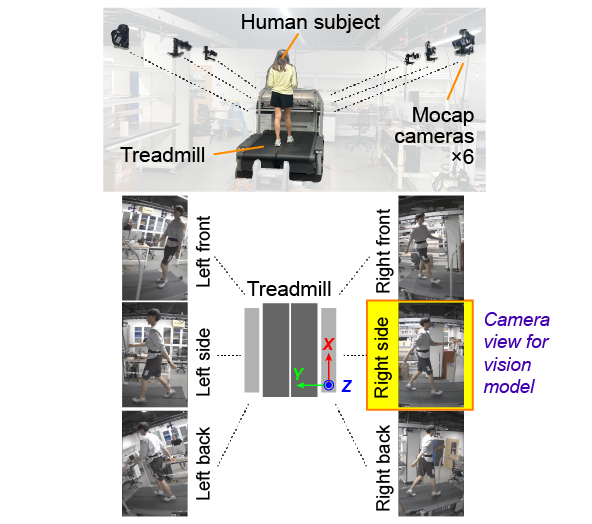}
        \caption{Motion capture setup for measuring ground-truth joint angles and vision model-calculated joint angles.}
        \label{Mocap_setup}
    \end{figure}

%% file: figures/tcn.tex
\begin{figure*}
    \centering
    \includegraphics[width=1.0\textwidth]{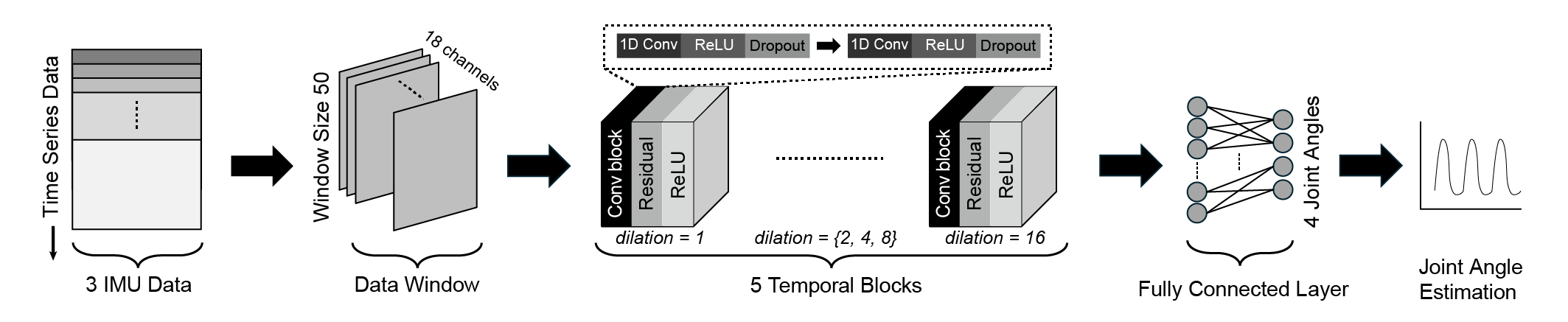}
    \caption{Structure of the Temporal Convolutional Network (TCN) for training the kinematic estimation model.}
    \label{tcn_model}
\end{figure*}

%% file: tables/vision_models_camera_table.tex
\begin{table}[ht]
    \caption{Joint Kinematics Estimation Errors of ViTPose Models Across Different Camera Views}
    \centering
    \renewcommand{\arraystretch}{1.2} 
    \begin{tabular}{l|cccccc}
        \textbf{RMSE {[}deg{]} \(\downarrow\)} & \multicolumn{6}{c}{\textbf{Camera view}} \\ 
        \cmidrule(lr){1-7}
        \shortstack{\textbf{Vision}\\ \textbf{model}} & 
        \shortstack{\textbf{Left}\\ \textbf{front}} & 
        \shortstack{\textbf{Left}\\ \textbf{side}} & 
        \shortstack{\textbf{Left}\\ \textbf{back}} & 
        \shortstack{\textbf{Right}\\ \textbf{front}} & 
        \shortstack{\textbf{Right}\\ \textbf{side}} & 
        \shortstack{\textbf{Right}\\ \textbf{back}} \\ 
        \cmidrule(lr){1-7}
        \textbf{ViTPose-small} & 16.30 & 10.01 & 10.88 & 12.75 & 9.66 & 10.82 \\
        \textbf{ViTPose-base}  & 16.50 & 10.03 & 10.74 & 13.04 & \cellcolor{yellow!50}\textbf{9.60} & 10.53 \\
        \textbf{ViTPose-large} & 16.46 & 10.07 & 10.90 & 13.64 & 9.79 & 10.41 \\
        \textbf{ViTPose-huge}  & 16.40 & 10.89 & 10.81 & 13.89 & 9.94 & 10.50 \\
    \end{tabular}
    \label{table_visionmodel}
\end{table}

%% file: figures/stiff_knee.tex
\begin{figure}
    \centering
    \includegraphics[width=0.5\textwidth]{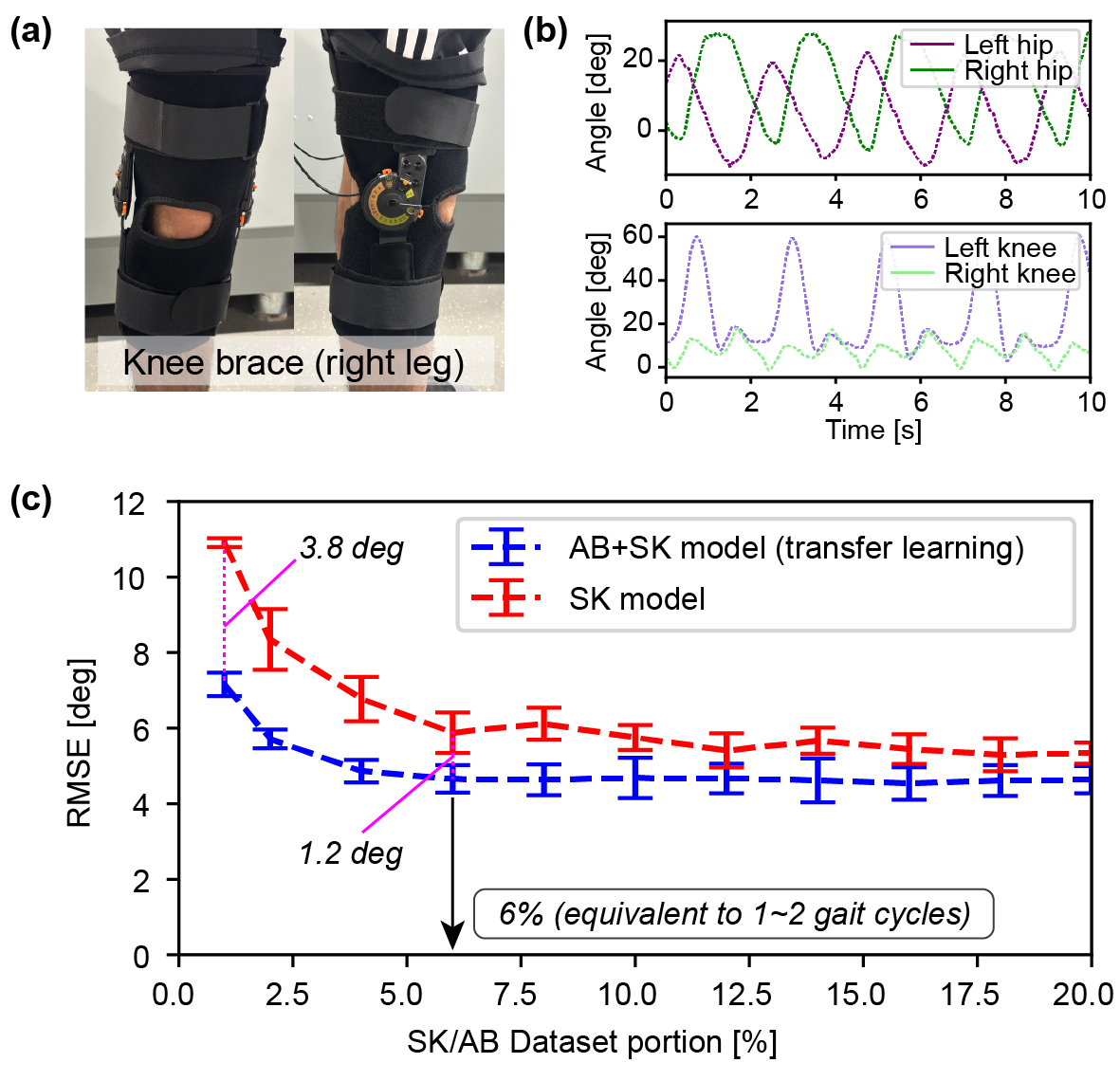}
    \caption{
        Experimental setup for evaluating stiff-knee (SK) behavior and the estimation error of a transfer-learned, adapted machine learning model.
        (a) Knee brace used to replicate the SK gait pattern.
        (b) Ground-truth joint angle trajectories during SK gait for a single subject walking at 1 m/s.
        (c) Comparison of estimation errors between the SK model and the adapted AB+SK model across varying SK-to-AB dataset ratios (mean ± SD, three SK subjects).}
    \label{Stiff_knee}
\end{figure}

%% file: figures/results.tex
\begin{figure}
    \centering
    \includegraphics[width=0.5\textwidth]{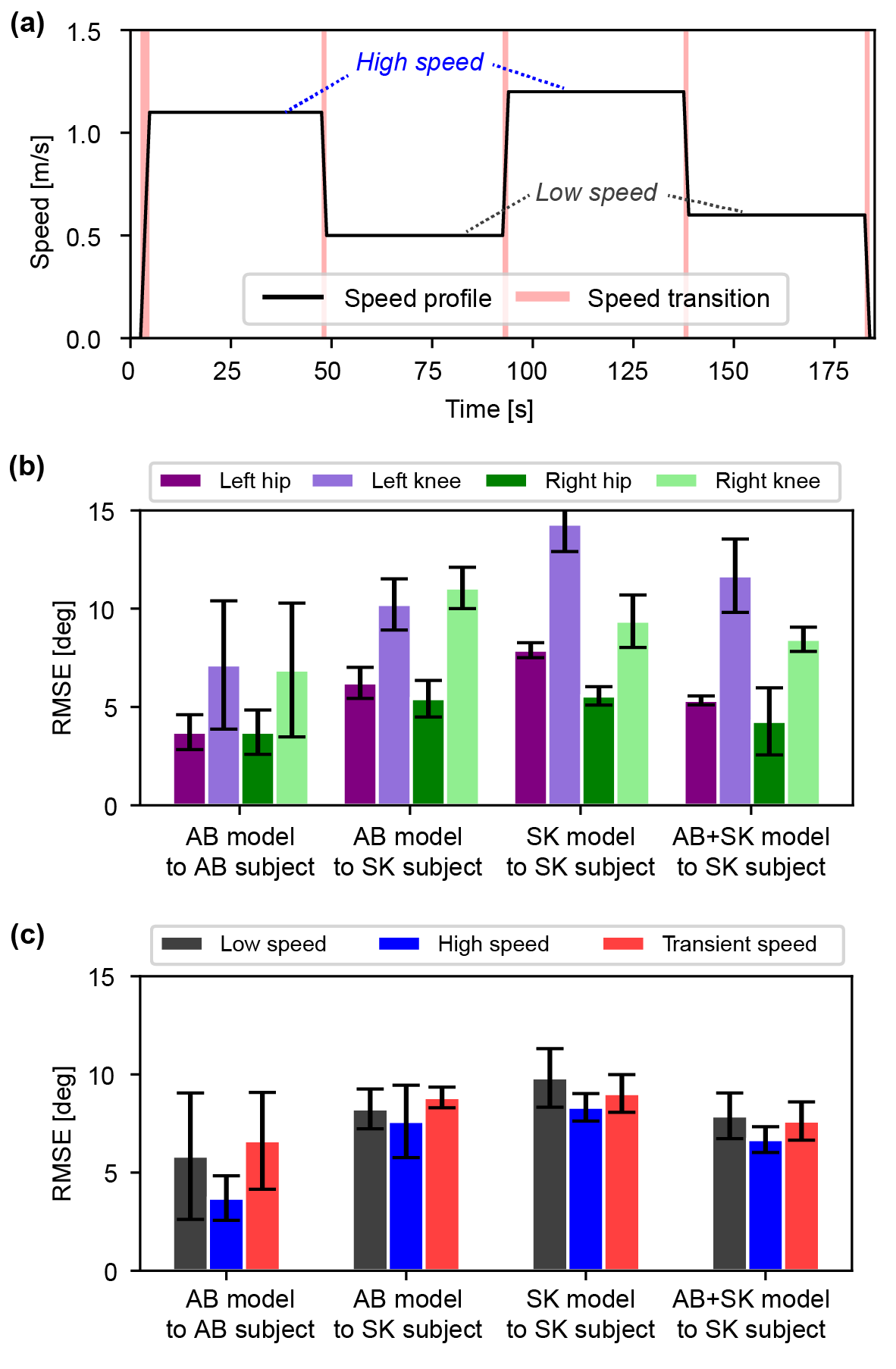}
    \caption{
    Results of the Validation Experiments on Real-Time Kinematic Estimation.
    (a) Speed profile of the treadmill used during the validation experiments.  
    (b) Estimation RMSE for each model across four lower-limb joints (mean ± SD, based on three SK subjects). The trial names indicate the specific TCN models (e.g., AB model, SK model, and AB+SK model) used to estimate data from specific subjects (e.g., AB subjects and SK subjects).  
    (c) Estimation RMSE for each model under varying treadmill speed conditions (mean ± SD, based on three SK subjects).}
    \label{Results}
\end{figure}

%% file: figures/results_joint.tex
\begin{figure}
    \centering
    \includegraphics[width=0.5\textwidth]{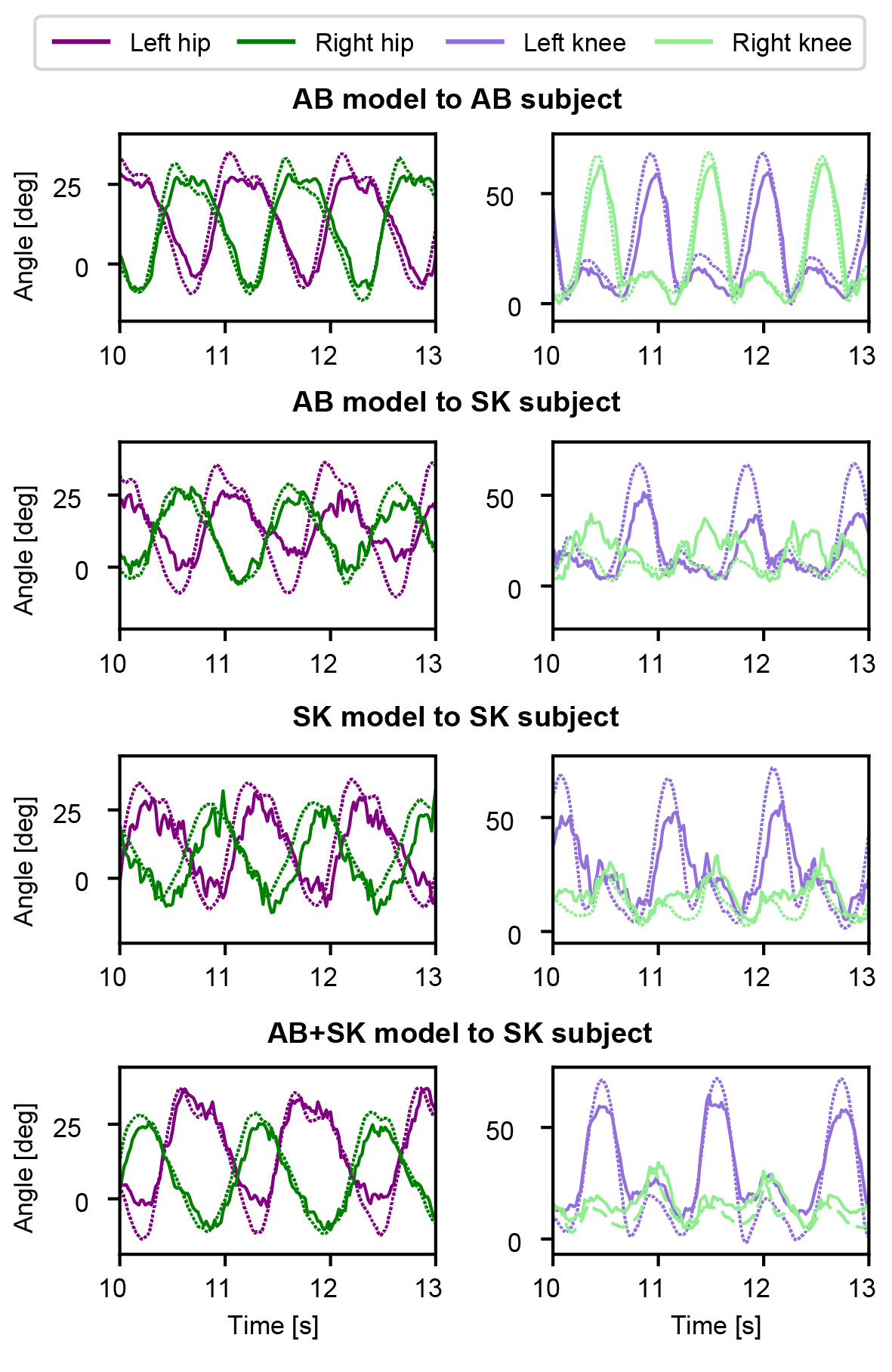}
    \caption{
        Exemplary plots of real-time kinematic estimations for each joint across experimental trials. The data corresponds to Subject 1 during a high-speed interval at 1.1 m/s. Solid lines represent the estimated joint angle values, while dotted lines indicate the ground-truth joint angle values.
        }
    \label{results_joint}
\end{figure}